\documentclass[sn-basic]{sn-jnl}

\usepackage[T1]{fontenc}
\usepackage[english]{babel}
\usepackage{amssymb}
\usepackage{amsmath}
\usepackage{mathtools}
\usepackage{paralist}
\usepackage{graphicx}
\usepackage{booktabs}
\usepackage{xcolor}
\usepackage{hyperref}
\usepackage{cleveref}

\DeclareMathOperator{\PD}{PartialDiameter}
\DeclareMathOperator{\ObsD}{ObservableDiameter}
\DeclareMathOperator{\E}{E}

\let\cref\Cref

\jyear{2023}%

\theoremstyle{thmstyleone}%
\newtheorem{theorem}{Theorem}
\newtheorem{corollary}[theorem]{Corollary}%

\theoremstyle{thmstyletwo}%

\theoremstyle{thmstylethree}%
\newtheorem{definition}{Definition}%

\begin{document}

\title{Selecting Features by their Resilience to the Curse of Dimensionality}


\author*[1]{\fnm{Maximilian} \sur{Stubbemann}}\email{stubbemann@cs.uni-kassel.de}

\author[1]{\fnm{Tobias} \sur{Hille}}\email{hille@cs.uni-kassel.de}

\author[1]{\fnm{Tom} \sur{Hanika}}\email{hanika@cs.uni-kassel.de}

\affil*[1]{\orgdiv{Knowledge \& Data
Engineering Group}, \orgname{University of Kassel, \city{Kassel}, \state{Hesse}, \country{Germany}}}


\abstract{Real-world datasets are often of high dimension and effected by the curse of
dimensionality. This hinders their comprehensibility and interpretability. To
reduce the complexity feature selection aims to identify features that are
crucial to learn from said data. While measures of relevance and pairwise
similarities are commonly used, the curse of dimensionality is rarely
incorporated into the process of selecting features. Here we step in with a
novel method that identifies the features that allow to discriminate data
subsets of different sizes. By adapting recent work on computing intrinsic
dimensionalities, our method is able to select the features that can discriminate
data and thus weaken the curse of dimensionality. Our experiments show that our method is
competitive and commonly outperforms established feature selection methods.
Furthermore, we propose an approximation that allows our method to scale to datasets consisting of millions
of data points. Our findings suggest that features that discriminate data and are
connected to a low intrinsic dimensionality are meaningful for learning procedures.}




\maketitle

\section{Introduction}

Contemporary datasets are of high dimension and their complexity continues to
increase. Thus, information derived by machine learning procedures is rarely
comprehensible or interpretable. This includes even procedures originally
categorized as explainable, such as decision trees. Addressing this problem, a
variety of works aims to simplify data through methods that reduce size or, in
particular, the dimensionality. An important class of such methods has the
emblematic name \emph{feature selection} (FS). By simply selecting to be
discarded feature dimensions, they preserve the explainability of the remaining
original dimensions, in contrast to approaches such as principal component
analysis. Although there are supervised and unsupervised types of FS, only
unsupervised procedures are suited to improve the understanding of data
regardless of a specific learning task. Moreover, they do not require costly
label information.

One important phenomenon which underlies the complexity of high dimensional data
is the \emph{curse of dimensionality}. The curse of dimensionality is a broad
term which is used for different phenomenons that arise in the context of
high-dimensional data. In this work, we follow the definition
of~\citet{pestov07}. According to Pestov, the curse of dimensionality describes
the situation where features are concentrated to specific regions and therefore
do not allow to discriminate different data points. Data that is strongly
affected by this phenomenon is considered to be of high \emph{intrinsic
dimension} (ID). Yet, in common feature selection methods, the curse of
dimensionality is rarely explicitly accounted for, even though it does a)
demonstrably influences learning success in high-dimensional real-world
scenarios, and b) potentially prevent common FS methods to choose the best
features to learn from. Thus, it is crucial to incorporate the intrinsic
dimension into the process of feature selection.

We meet this challenge by proposing an unsupervised feature selection method
that picks features based on their measurable ability to discriminate different
data points. While other methods are based on feature relevance measures, such
as variances, or select features by discarding strongly correlated ones, our
method ranks features by their resilience against the
curse of dimensionality in the sense of Pestov. For this, we build on recent work~\citep{hanika22,
stubbemann23} which studies intrinsic dimensionality for \emph{geometric
datasets}. By adapting corresponding ideas for computing the intrinsic
dimensionality, we derive an algorithm that allows to find precisely those
features that are able to tame the influence of the curse of dimensionality. Furthermore,
adapting the speed-up techniques proposed in~\citet{stubbemann23} will allow feature selection in settings with large-scale
datasets comprised of millions of data points. While \citep{hanika22} and
\citep{stubbemann23} are focused on
computing the intrinsic dimensionality of datasets, we will rank and select
individual features by their ability to discriminate data points. Thus, we identify
features that are harmed by the curse of dimensionality by a comparable low extent.

We experimentally show on real-world datasets that features selected
by our proposed method are meaningful to learn from. To be more
specific, we experiment on the \emph{OpenML-CC18 Curated
  Classification benchmark} (Open18)~\citep{bischl19} and witness that
our method is competitive or even outperforms established feature
selection methods with respect to feature selection for classification
tasks. Furthermore, we use the \emph{Open Graph Benchmark}~\citep{hu20}
to show that our method is capable of feature selection for
large-scale learning with \emph{Graph Neural Networks}. To sum up, our
experiments indicate that features that are able to discriminate data and thus
weaken the curse of dimensionality are highly relevant for learning tasks. Our
code is publicly available.\footnote{\url{https://github.com/mstubbemann/FSCOD}}

\section{Related Work}
We focus on unsupervised feature selection, which is an established research
topic~\citep{fernandez20, cai18,alelyani18}. Established methods are often
focused on the identification of outstanding features by identifying the
relevant features for clustering~\citep{dy04, breaban11, dutta14}. Another
directions focuses on selecting a subset of important and dissimilar
features~\citep{zhao07,ferreira12, mitra02, hanika19}. Here, feature importance can for
example be measured via variances and similarities via correlation coefficients.
While applying such methods may result in a decreasing intrinsic dimensionality, they
do not explicitly account for the curse of dimensionality in the selection
process.

A small amount of research has incorporated intrinsic
dimension estimators into feature selection~\citep{gomez10, golay17, mo12,
faloutsos10}. However, these method either only use the ID to determine the
amount of features to select~\citep{gomez10} or they are
based on recomputing intrinsic dimensionalities after discarding features which
limits scalability~\citep{golay17, mo12}. Even if applicable to larger datasets,
these methods build on notions of intrinsic dimensionality that do not aim to
quantify the curse of dimensionality~\citep{faloutsos10}. Instead, these notions
work under the assumption that the data lies on a manifold of lower Euclidean
dimension. The goal of the ID notion is then to approximate the dimension of
this manifold.

In contrast, we build our feature selection procedure on a notion of ID which quantifies
the influence of the curse of dimensionality which occurs when data points can
not be discriminated. In the works of Pestov~\citep{pestov00,
pestov07, pestov08}, all $1-$Lipschitz functions are considered
as potential features to discriminate data, which often hinders the
practical computation. This was tackled by defining intrinsic
dimensionality for \emph{geometric datasets}~\citep{hanika22}, where a set of
features to consider can be defined beforehand. The computation and
approximation on large-scale real world data has recently been made
possible~\citep{stubbemann23}. We use these findings to incorporate Pestovs
notions of the curse of dimensionality in the process of unsupervised feature
selection. Note, that~\citet{hanika22} and~\citet{stubbemann23} were solely focused on the intrinsic dimensionality
of datasets and thus of the question how all features together can discriminate
data points. In contrast, we present a novel approach to rank and select
individual features by their ability to discriminate data points.

\section{Feature Selection via Discriminability}
\label{sec:fs}

We present our work in a general setting for datasets, i.e., for
\emph{geometric datasets} $\mathcal{D}=(X,F,\mu)$~\citep{hanika22},
where $X$ is a set of \emph{data points} and $F \subseteq
{\mathbb{R}}^{X}$ is a set of \emph{feature functions} from $X$ to
$\mathbb{R}$. We require  $\sup_{x,y\in X} d_F(x,y) < \infty$,
where $d_F(x,y) \coloneqq \sup_{f \in F}\lvert f(x)- f(y)\rvert$. 
Furthermore, $(X,d_F)$ has to be a complete and
separable metric space with $\mu$ being a Borel probability measure on
$(X,d_F)$.

From this point on, we consider the special case of \emph{finite
geometric datasets}, i.e., $0 <\lvert X\rvert ,\lvert F\rvert  <\infty$,  with $\mu$ being the normalized counting
measure. \citet{hanika22}
derived an axiomatization for functions to be \emph{dimension
  functions} of geometric datasets. This axiomatization is based on the concentration of
measure phenomenon~\citep{gromov83,milman88,milman00} and its linkage
to the curse of dimensionality by
Pestov~\citep{pestov00,pestov07,pestov08,pestov10}. The latter defines the
curse of dimensionality as the phenomenon of features concentrating
near their means or medians and not being able to discriminate
data. To quantify this phenomenon the \emph{partial
  diameter} is used, which determines to which extent a feature
can discriminate sets of a specific measure $\alpha$. For finite geometric
datasets the partial diameter has the form
\begin{equation*}
  \PD(f, 1 - \alpha)_{\mathcal{D}} = \mathop{\min_{M
  \subseteq X}}_{\lvert M\rvert =c_{\alpha}}\max_{x,y \in M}\lvert f(x) - f(y)\rvert ,
\end{equation*}
where $c_\alpha \coloneqq \lceil \lvert X  \rvert\lvert(1-\alpha) \rceil$. Based on this, the \emph{observable
diameter} reflects how the feature set $F$ can discriminate data of a
specific measure. More formally, $\ObsD(\mathcal{D}, -\alpha) \coloneqq
\sup_{f \in F} \PD(f, 1 - \alpha)_{\mathcal{D}}$. This notions allows to determine for
a geometric dataset $\mathcal{D}=(X,F,\mu)$ the ability of $F$ to discriminate
the data points in $X$. This is done by considering the Observable
Diameter over all possible values for $\alpha$ via
\begin{equation*}
  \Delta(\mathcal{D}) \coloneqq \int_{0}^{1} \ObsD(\mathcal{D}, -\alpha) d\alpha,
\end{equation*}
which has in the case of finite geometric datasets the form
\begin{equation}
  \label{eq:delta_long}
      \Delta(\mathcal{D}) = \frac{1}{\lvert X\rvert }\sum_{k=2}^{\lvert X\rvert }\max_{f \in F}\mathop{\min_{M
        \subseteq X}}_{\lvert M\rvert =k}\max_{x,y \in M}\lvert f(x) - f(y)\rvert,
\end{equation}
as shown in~\citet{stubbemann23}. With the usage of the notation  $\phi_{k,f}\coloneqq \min_{M \subseteq
X,\lvert M\rvert =k}\max_{x,y \in M}\lvert f(x) - f(y)\rvert $, and $\phi_k \coloneqq \max_{f \in F}
\phi_{k,f}$ the discriminability $\Delta(\mathcal{D})$ can be
rewritten as
\begin{equation}
  \label{eq:delta}
  \Delta(\mathcal{D})=\frac{1}{\lvert X\rvert }\sum_{k=2}^{\lvert X\rvert } \max_{f
\in F}\phi_{k,f}= \frac{1}{\lvert X\rvert }\sum_{k=2}^{\lvert X\rvert } \phi_k.
\end{equation} 
In the following, we consider feature selection with a fixed feature budget in
an unsupervised setting, i.e., without any known labels of a specific
classification task at selecting time.

\paragraph{Problem Statement.}
Given a finite geometric dataset $\mathcal{D}=(X,F, \mu)$ and a natural
number $n_F \ll \lvert F\rvert $, select the $n_F$ most important features of $F$. 

\subsection{Selection of Discriminating Features}
\label{sec:crf}
The definition of discriminability as in~\cref{eq:delta_long}
and~\cref{eq:delta} quantifies to which extent the set of all features can
discriminate data subsets of different cardinality. The main idea of our feature
selection algorithm is to rank features by their ability to discriminate data
subsets of different cardinality by solely using this feature. 

\begin{definition}[Discriminability]
  The \emph{discriminability of $\mathcal{D}$ with respect to feature $f
  \in F$} is defined as
  \begin{equation}
    \label{eq:f}
    {\Delta(\mathcal{D})_f^{*}} \coloneqq \frac{1}{\lvert X \rvert} \sum_{k=2}^{\lvert X \rvert} \mathop{\min_{M
    \subseteq X}}_{\lvert M\rvert =k} \max_{x,y \in M} \lvert f(x) - f(y) \rvert = \frac{1}{\lvert X \rvert}\sum_{k=2}^{\lvert X \rvert} \phi_{k,f}.
  \end{equation}  
\end{definition}

\paragraph{Enhancing Robustness against Outliers}
Note, that one data point with an outstanding value $f(x)$ can have a strong
influence on ${\Delta(\mathcal{D})_f}^{*}$ via drastically increasing
$\phi_{\lvert X \rvert,f}$. To weaken this phenomenon, we propose to weight
$\phi_{k,f}$ higher for smaller values of $k$. This leads to the following
definition.

\begin{definition}[Normalized Discriminability and Normalized Intrinsic Dimensionality]
  The \emph{normalized discriminability of $\mathcal{D}$ with respect to $f$} which
  we define as
  \begin{equation}
    \label{eq:normalized_f}
    \Delta(\mathcal{D})_f \coloneqq \frac{1}{\lvert X \rvert}\sum_{k=2}^{\lvert X \rvert} \frac{1}{k} \phi_{k,f}.
  \end{equation} 
  The \emph{normalized intrinsic dimensionality of $\mathcal{D}$ with respect to
  $f$} is then given via
  \begin{equation}
    \label{eq:id}
    \partial (\mathcal{D})_f \coloneqq \frac{1}{\Delta(\mathcal{D})_f^{2}}.
  \end{equation} 
\end{definition}

We then can rank features by their discriminability and select the features with
the highest discriminability/ lowest intrinsic dimensionality. We call the
resulting method \emph{features selection via discriminability} (FSD). It is
depicted in~\cref{algo1}.

\begin{algorithm}[t]
  \caption{Feature Selection via Discriminability(FSD)}\label{algo1}
\begin{algorithmic}[1]
\Require Finite geometric dataset $\mathcal{D}=(X, F, \mu)$. Natural number $k <<
\lvert F \rvert$ of features to select from  $F$.
\For{$f \in F$}
    \State $\Delta(\mathcal{D})_f \coloneqq \frac{1}{\lvert X \rvert}\sum_{k=2}^{\lvert X \rvert} \frac{1}{k} \phi_{k,f}$
    \State $\partial(\mathcal{D})_f=\frac{1}{\Delta(\mathcal{D})_f^{2}}$
\EndFor
\State Set $l_F$ as the list of all $f \in F$ ordered by ascending
$\partial(\mathcal{D})_f$.
\State \Return $l_F[:k]$
\end{algorithmic}
\end{algorithm}
\subsection{Discarding Highly Correlated Features}
\label{sec:discarding}
Our methods select features by their ability to separate data points.
However, it does not consider connections between individual features, as for
example correlations. Thus, if for example the two features that separate the
data set are nearly identical our method may select both as though selecting
one of these features would be sufficient as the second one gives no extra
information. To prevent such cases, we propose to incorporate correlation
coefficients into our feature selection  process, if desired.
In order to do so, we define an additional number $n_c$ of features to remove
via correlation coefficients. We then iteratively discard one of the two features
$f_1,f_2$ with the maximal pearson correlation coefficient. The version of our
algorithm that incorporates this preprocessing is called \emph{Feature Selection
via Discriminability and Correlation} (FSDC).

\section{Feature Selection for Large-Scale Data}

Since computing $\phi_{k,f}$ is in $\mathcal{O}(\lvert X \rvert-k)$ for fixed $f
\in F$~\citep{stubbemann23}, computing $\Delta(\mathcal{D})_f $ is in
$\mathcal{O}( \sum_{k=2}^{\lvert X \rvert}\lvert X \rvert-k)=\mathcal{O}(\lvert
X \rvert )$. Hence, we have a worst case runtime which scales quadratic with
$\lvert X\rvert$. Thus, FSD(C) is applicable to medium-sized data sets
with thousands of data points. However, it is not tailored to large-scale data
with millions of data points.

Earlier work~\citep{stubbemann23} has used
so called \emph{support sequences}, i.e. strictly increasing finite sequences of
the form $s=(2=s_1,\dots,s_l=\lvert X \rvert)$ to approximate
$\Delta(\mathcal{D})$ and $\partial (\mathcal{D})$ by only computing
$\phi_{s_i,f}$ for $s_i \in s$.

Inspired by this, we approximate the discriminability and intrinsic
dimensionality with respect to a feature $f \in F$. The approximation
is based on the result, that the map $k \mapsto \phi_{k,f}$ is monotonically
increasing~\citep{stubbemann23}. Using this result, we replace for $s_i < j <
s_{i+1}$ the value $\phi_{j,f}$ by $\phi_{s_i,f}$ or $\phi_{s_{i+1}, f}$. More
formally, we get the following definition.
\begin{definition}[Upper and Lower Normalized Discriminability]
  For a feature $f \in F$ and a support sequence $s$ we call
  \begin{equation*}
    \Delta(\mathcal{D})_{s,f}^{+}\coloneqq \frac{1}{\lvert X \rvert}
    \left(\sum_{i=1}^l \frac{1}{s_i}\phi_{s_i, f} +
    \sum_{i=1}^{l-1}\sum_{s_i<j<s_{i+1}} \frac{1}{j}\phi_{s_{i+1}, f}\right)
  \end{equation*}
  the \emph{upper normalized discriminability with respect to $f$ and $s$} and 
  \begin{equation*}
    \Delta(\mathcal{D})_{s,f}^{-}\coloneqq \frac{1}{\lvert X \rvert}
    \left(\sum_{i=1}^l \frac{1}{s_i}\phi_{s_i, f} +
    \sum_{i=1}^{l-1}\sum_{s_i<j<s_{i+1}} \frac{1}{j}\phi_{s_{i}, f}\right)
  \end{equation*}
  he \emph{lower normalized discriminability with respect to $f$ and $s$}.
\end{definition}

\begin{definition}[Upper, Lower and Approximated Normalized Intrinsic Dimensionality]
  We define the \emph{upper/lower normalized intrinsic dimensionality
with respect to $f$ and $s$} via $\partial (\mathcal{D})_{s,f}^{+} \coloneqq
\frac{1}{\left(\Delta(\mathcal{D})_{s,g}^{-}\right)^{2}}$ and $\partial (\mathcal{D})_{s,f}^{-} \coloneqq
\frac{1}{\left(\Delta(\mathcal{D})_{s,f}^{+}\right)^{2}}$. We then rank the features in
descending order by their \emph{ approximated normalized intrinsic dimensionality
with respect to $f$ and $s$} which is defined via
\begin{equation}
  \label{eq:app_id}
  \partial({\mathcal{D}})_{s,f} \coloneqq \frac{\partial(\mathcal{D})_{s,f}^{+} + \partial(\mathcal{D})_{s,f}^{-}}{2}.
\end{equation}
\end{definition}

The resulting algorithm for large-scale datasets is depicted in~\cref{algo2}. We
rank features via ascending $ \partial({\mathcal{D}})_{s,f}$. If we want to
select $k$ features, we choose
the first $k$ features of the resulting order.

\begin{algorithm}[t]
  \caption{Large-Scale Feature Selection via Discriminability(LSFSD)}\label{algo2}
\begin{algorithmic}[1]
\Require Finite geometric dataset $\mathcal{D}=(X, F, \mu)$. Natural number $k <<
\lvert F \rvert$ of features to select from  $F$. Support sequence
$s=(2=s_1,\dots,s_l=\lvert X \rvert)$.
\For{$f \in F$}
    \State $   \Delta(\mathcal{D})_{s,f}^{+}\coloneqq \frac{1}{\lvert X \rvert}
    \left(\sum_{i=1}^l \frac{1}{s_i}\phi_{s_i, f} +
    \sum_{i=1}^{l-1}\sum_{s_i<j<s_{i+1}} \frac{1}{j}\phi_{s_{i+1}, f}\right)$
    \State     $\Delta(\mathcal{D})_{s,f}^{-}\coloneqq \frac{1}{\lvert X \rvert}
    \left(\sum_{i=1}^l \frac{1}{s_i}\phi_{s_i, f}+
\sum_{i=1}^{l-1}\sum_{s_i<j<s_{i+1}} \frac{1}{j}\phi_{s_{i},
f}\right)$
    \State $\partial (\mathcal{D})_{s,f}^{+} \coloneqq
    \frac{1}{\left(\Delta(\mathcal{D})_{s,g}^{-}\right)^{2}}$ 
    \State $\partial (\mathcal{D})_{s,f}^{-} \coloneqq
    \frac{1}{\left(\Delta(\mathcal{D})_{s,g}^{+}\right)^{2}}$ 
\State $\partial({\mathcal{D}})_{s,f} \coloneqq
\frac{\partial(\mathcal{D})_{s,f}^{+} + \partial(\mathcal{D})_{s,f}^{-}}{2}$
    
\EndFor
\State Set $l_F$ as the list of all $f \in F$ ordered by ascending
$\partial(\mathcal{D})_{s,f}$.
\State \Return $l_F[:k]$
\end{algorithmic}
\end{algorithm}

\subsection{Error Ratios of Approximations}
Note, that $\partial({\mathcal{D}})_{s,f}$ as defined in~\cref{eq:app_id} only gives
an approximation of $\partial({\mathcal{D}})_{f}$ as defined in~\cref{eq:id}.
Thus ranking the features by $\partial({\mathcal{D}})_{s,f}$ instead of
$\partial({\mathcal{D}})_{f}$ may lead to a different ordering. We are
interested in the amount of such changes. Let $l_s
\coloneqq (f_{i_1}, \dots, f_{i_{\lvert F \rvert}})$ be the list of the features
in $F$ ordered by ascending $\partial({\mathcal{D}})_{s,f}$.

\begin{definition}[Error Ratio]
  The \emph{error
ratio of $s$} is then given by
\begin{equation}
  \label{eg:real_errors}
  \E(s)^{*} \coloneqq \frac{2\lvert \{(k,l) \in \{1,\dots, \lvert F \rvert\} \mid k <l \land \partial({\mathcal{D}})_{f_{i_k}} > \partial({\mathcal{D}})_{f_{i_l}}\}\rvert}
  {\lvert F \rvert (\lvert F \rvert - 1)}.
\end{equation}
\end{definition}

Computing $\partial(\mathcal{D})_{s,f}$ for all $f \in F$ is especially of
interest when computing $\partial(\mathcal{D})_{f}$ is not feasible. In such
circumstances, it is also not possible to compute $\E(s)^{*}$. Hence, we need an
approximation or upper bound of $\E(s)^{*}$ which can be computed without
needing $\partial(\mathcal{D})_{f}$. 

\begin{definition}[Maximal Error Ratio]
We define the \emph{maximal error ratio} of $s$ via 
\begin{equation}
  \label{eq:max_errors}
  \E(s) \coloneqq \frac{2 \lvert \{(k,l) \in \{1,\dots, \lvert F \rvert\} \mid k <l \land \partial({\mathcal{D}})_{s,f_{i_k}}^{+} > \partial({\mathcal{D}})_{s,f_{i_l}}^{-}\}\rvert }
  {\lvert F \rvert (\lvert F \rvert - 1)}.
\end{equation}
\end{definition}
The maximal error ratio $E(s)$ can be computed without knowing $\partial(\mathcal{D})_f$ for
all $f \in F$.   Per definition it holds that
$\partial(\mathcal{D})_{s,f}^{-} \leq \partial(\mathcal{D})_{s,f} \leq
\partial(\mathcal{D})_{s,f}^{+}$. Thus, for given features $f,g \in F$ with
$\partial(\mathcal{D})_{s,f}^{+} \leq \partial(\mathcal{D})_{s,g}^{-}$ we can
conclude $\partial(\mathcal{D})_{f} \leq \partial(\mathcal{D})_{s,f}^{+} \leq
\partial(\mathcal{D})_{s,g}^{-} \leq \partial(\mathcal{D})_{g}$. Thus, we get
the following corollary.

\begin{corollary}
 For each support sequence $s$ it holds that
  \begin{equation}
    \label{eq:bound}
    \E(s)^{*} \leq \E(s).
  \end{equation}
\end{corollary}

To sum up, we now have an approximation algorithm which allow us to rank the
features by their intrinsic dimensionality and we can efficiently bound the
amount of errors this approximation produces. We denominate this method with
\emph{large-scale feature selection via discriminability (and correlation)}(LSFSD(C)).

\section{Experiments}
In the following we empirically examine the following hypothesis. \emph{Discriminative
features are meaningful features with respect to learning performances}. The aim
is \textbf{not} to develop a new feature selection procedure that surpasses all
established methods.
Thus, we do not compare with all state-of unsupervised feature selection
procedures. Instead we use a small set of representative baselines that consists of
established methods and  common-sense baselines based on correlation and variances.
Furthermore, we compare our approach to random feature selection to evaluate if
our features are indeed meaningful with respect to classification. 

In all our experiments, we consider the features $F$ to be the coordinate
projections, i.e., the feature functions are given via the data columns. We
evaluate our feature selection method regarding to classification performances.
We experiment with a Logistic Regression classifier on the \emph{OpenML-CC18
Curated Classification benchmark}
(Open18)\footnote{\url{https://www.openml.org/search?type=study&study_type=task&id=99&sort=tasks_included}}
~\citep{bischl19}. To further evaluate our selection procedure on modern
large-scale data we also select features for classification with \emph{Graph
Neural Networks} (GNNs) on a subset of the \emph{Open Graph
Benchmark~}\citep{hu20}. In both experiments, we evaluate how to classify on a
small subset of the features. To be more detailed, we only want to keep 10\% of all
features. We evaluate our feature selection with and without the preprocessing
procedure of~\cref{sec:discarding}. In the case with the preprocessing step, we
use it do discard 10\% of the features. In both experiments, we will also report
results for classification on the full feature set. We evaluate against the
following baselines.
\begin{itemize}
\item\textbf{Random.} Randomly selecting the 10\% of features.

\item\textbf{Correlation Based.} Only use the correlation based feature selection
as proposed in~\cref{sec:discarding} until only 10\% of the features are left.

\item\textbf{Variance Based.} Select the 10\% of the features with the highest
variance. We choose this baseline because it is a straight forward approach to
for selecting features by their individual importance.

\item\textbf{SPEC.\citep{zhao07}} SPEC first builds a complete graph between the
individual data points with edge weights indicating similarities of the points
and then uses spectral graph theory for estimating feature relevance. In our
experiments, we use an RBF Kernel for similarity and evaluate feature relevance
via $\phi_2$. For details, we refer to \citet{zhao07}. We choose this
baseline because it was the univariate filter method with the highest classification
accuracy in a recent survey on unsupervised feature
selection~\citep{fernandez20}.

\item\textbf{RRFS.\citep{ferreira12}} RRFS combines a function @sim which measures
similarity of feature pairs and a relevance measure @rel that evaluates
the importance of features. It iteratively chooses the next most
relevant feature that has a similarity to the last chosen feature which below a specific threshold $t$. For comparison, we choose $t$ to be the
correlation coefficient value of the last pair of features that was used to
discard a feature via \cref{sec:discarding}.
We use the Pearson correlation coefficient for similarity evaluation and
evaluate the relevance of features via their variance. For more details, we
refer to~\citet{ferreira12}. We use this baseline because it was the
multivariate filter method with the highest classification accuracy in the survey
mentioned above~\citep{fernandez20}.

\end{itemize}

\subsection{Feature Selection for Logistic Regression}
\begin{table}[t]
\caption{Results on \textbf{Open18} experiment. We report the accuracies
for our method with (FSDC) and without preprocessing (FSD) via correlation and
for all baselines}
\begin{center}
  \scalebox{.75}{
  \begin{tabular}{|cl|c|c|cc|ccccc|}
    \toprule
Task ID &Dataset& $\lvert F \rvert$& Full & FSDC& FSD  & Random &Corr &Vari & RRFS& SPEC\\
\midrule
14 & mfeat-fourier&76& .813&.694&\textbf{.753}&.433$\pm$.032 &.457&.751&.751&.630\\
16 & mfeat-karh.&64 & .958 & .695 & \textbf{.822} & .466$\pm$.028 & .288 & .810
& .810 & .680\\
45& splice & 60 &.918 & .531 & .611 & .580$\pm$.026 & .534 & .725 & .647 &
\textbf{.806 }\\
9910 & Bioresponse& 1776 & .754 & .745&\textbf{.751}&.704$\pm$.008 & .739 & .747 &
.707 & .550\\
9977 & nomao & 118&.947 & \textbf{.907} & \textbf{.907} & .835$\pm$.012 & .752 & .793 &
.793 & .759 \\
9981& cnae-9 & 856&.944 & .884 & .884 & .391$\pm$.032& .404 & \textbf{.888} & .144
& .128 \\
9985 & 1st-order&51 & .479 & \textbf{.434} & \textbf{.434} & .420$\pm$.002& .416
&.419 & .419& .418\\
167125 & Internet-Advertisment & 1558 & .971 & .962 & \textbf{.963}& .936$\pm$.003 & .899
& \textbf{.963} & \textbf{.963} & .877 \\
\bottomrule
  \end{tabular}}
\end{center}
  \vspace{.3cm}

  \label{tab:open18}
\end{table}

We select features as explained above and then use a Logistic Regression
classifier. We use the data splits that are provided by
Open18 and report mean test accuracies over all splits. We report means
over 10 runs of this experiment. Note, that the Logistic Regression of Scikit
Learn~\citep{pedregosa11} using the default solver leads to deterministic
results. As only the random selection method is non-deterministic, this
is the only baseline where results of different repetitions vary.

We only experiment on a subset of \textbf{Open18}. To be more detailed, we discard all
datasets with NaN values or only binary features. Furthermore, we discard all
datasets where the Logistic Regression classifier of
Scikit-Learn was not always able to converge on the full
feature set. We use default
parameters with the exception that we changed the maximal iterations from $100$ to
$1000$ to increase the chance of convergence. Finally, we arrive at $8$ out of $72$
datasets of \textbf{Open18}. The results are depicted in~\cref{tab:open18}.

\subsubsection{Results and Discussion.}
For all datasets, FSD outcompetes the random baseline and for 7 out of 8
datasets FSDC outperforms random selection. Hence, selecting features by their
discriminability a reasonable aproach to identify relevant features for
learning. Furthermore, for $6$ out of $8$ baselines, FSD surpasses all
baselines. This indicates that our selection approach is competetive with
established feature selection methods.

Adding the correlation-based feature selection to FSD does not increase performances and feature selection solely based on
correlation coefficients leads to comparable low accuracies. Furthermore, RRFS,
which, in our configuration, combines feature variances with dropping strongly
correlated features does not surpass the selection of features solely based on
variance. All this indicates that the incorporation of correlation between
features is not useful in this scenario. It also stands out, that FSDC sometimes
lead to the same accuracy than FSD. In these cases it stands to reason that FSD
does not select any features that are dropped by FSDC as preprocessing.

Overall, selecting only 10\% often lead to a high drop compared to the original
accuracy, indicating that such a small feature budget often not allows for
sufficient feature selection. Thus, our experiment mainly gives insights in cases
where feature selection is done because only using a small subset of features is
computational feasible or to get new insights into the data and
classification behavior. It is primarily not designed for feature selection to
enhance classification accuracy. Here, the amount of selected features should be
set higher. However, the competitiveness of our approach supports our hypothesis
that discriminability of features is connected to their relevance for learning.

\subsection{Feature Selection for Graph Neural Networks}
\label{sec:exp_gnn}

\begin{table}[t]
  \caption{Results of our experiments on OGBN. We report, the accuracies for our method  with (LSFSDC) and
  without (LSFSD) and for all baselines except of SPEC.}
  \label{tab:ogbn}
 \begin{center}
  \scalebox{.74}{
 \begin{tabular}{|c|c|c|cc|cccc|}

    \toprule

Dataset& $\lvert F \rvert$& Full & LSFSDC& LSFSD  & Random
&Corr &Vari & RRFS\\
\midrule
arxiv& 128& $.691 \pm .006$&$.536 \pm .005$& $\textbf{.543} \pm .005$ &$.500 \pm .021$ & $.461 \pm .006$
&$.543 \pm .008$&$.534 \pm .003$\\
products&100 & $.748 \pm .006$ & $.529 \pm .007$& $\textbf{.545} \pm .005$ & $.413 \pm.038$
& $.395 \pm .006$ & $.485 \pm .006$ & $.519 \pm .008$ \\
mag&128 & $.387 \pm .004$ & $\textbf{.292} \pm .005$ & $.283 \pm .005 $& $.274 \pm .006$
& $.269 \pm .004$ & $.292 \pm .002 $& $.282 \pm .004$\\
\bottomrule
  \end{tabular}}
\end{center}
  \vspace{.3cm}
\end{table}

We now evaluate to which extent our feature selection methods
helps to select features for training Graph Neural Networks. 
For this, we use \emph{ogbn-arxiv}, \emph{ogbn-mag} and
\emph{ogbn-products}\footnote{\url{https://ogb.stanford.edu/docs/nodeprop/}}
from Open Graph Benchmark~\citep{hu20}. These datasets are fundamentally larger
then the ones from \textbf{Open18}, with a nodeset size of $169,343$
(\textbf{ogbn-arxiv}), $2,449,029$ (\textbf{ogbn-products}) and $1,939,743$
\textbf{(ogbn-mag)}. 

To use LSFD(C) we need to choose a support sequence. As in~\citet{stubbemann23},
we use log scale spacing. For this, we first choose a geometric sequence
$\hat{s}=(s_1,\dots s_l)$ of length $l=10,000$ with $s_1 = \lvert X \rvert$ and
$s_l=2$ and use the support sequence $s$ which results from $s'=(\lfloor \lvert
X \rvert +2 - s_1 \rfloor, \dots,
\lfloor \lvert X \rvert + 2 - s_l \rfloor)$ via discarding duplicated elements.

For training and classification, we use a plain SIGN model~\citep{rossi20} with
one hidden-layer of dimension $512$ and do $2-$hop neighborhood aggregation. We
train with an Adam optimizer with learning rate of $0.001$ and weight decay of
$0.0001$. We train for a maximum of $1000$ epochs with a patience of $15$ epochs
with respect to validation accuracy. We use a batch size of $256$. We dropout at
the input layer with a probability of $0.1, 0.3, 0.5$ and at the hidden layer
with probability of $0.5, 0.4, 0.5$ for \textbf{ogbn-arxiv},
\textbf{ogbn-products} and \textbf{ogbn-mag}, respectively.

We report test accuracies, displaying
means and standard deviations over $10$ rounds. We use all methods and baselines
reported above. For SPEC, computation of the RBF kernel was not
possible on our Server with $125$ GB RAM due to memory costs. This was true for
using Scikit-Learn as well with plain numpy. The results are depicted in~\cref{tab:ogbn}.

\subsubsection{Results and Discussion.}
In this experiment, preprocessing via~\cref{sec:discarding} indeed can help as LSFSDC
lead to higher accuracies then using LSFSD for \textbf{ogbn-mag}. LSFSDC also surpasses
feature selection based solely on correlations. Note, that RRFS, the
other method that combines similarities between features and selection by some
measure of feature relevance, do lead to fundamentally worse results. Thus,
one can not follow that combining similarity and relevance is the general key to
successful feature selection. This is supported by the fact, that feature
selection solely based on feature variance leads to high accuracies.
For \textbf{ogbn-products} and \textbf{ogbn-arxiv}, LFSD lead to the best performance, for
\textbf{ogbn-mag}, the highest accuracy is reached with LSFSDC.
In this experiment, our selection method is again competitive with the baselines and surpasses
random selection. This supports the hypothesis that selecting features via their
discriminability is meaningful. 

\section{Parameter Study on Maximal Errors}

\begin{figure}[t]
  \centering
  \includegraphics[width=\linewidth]{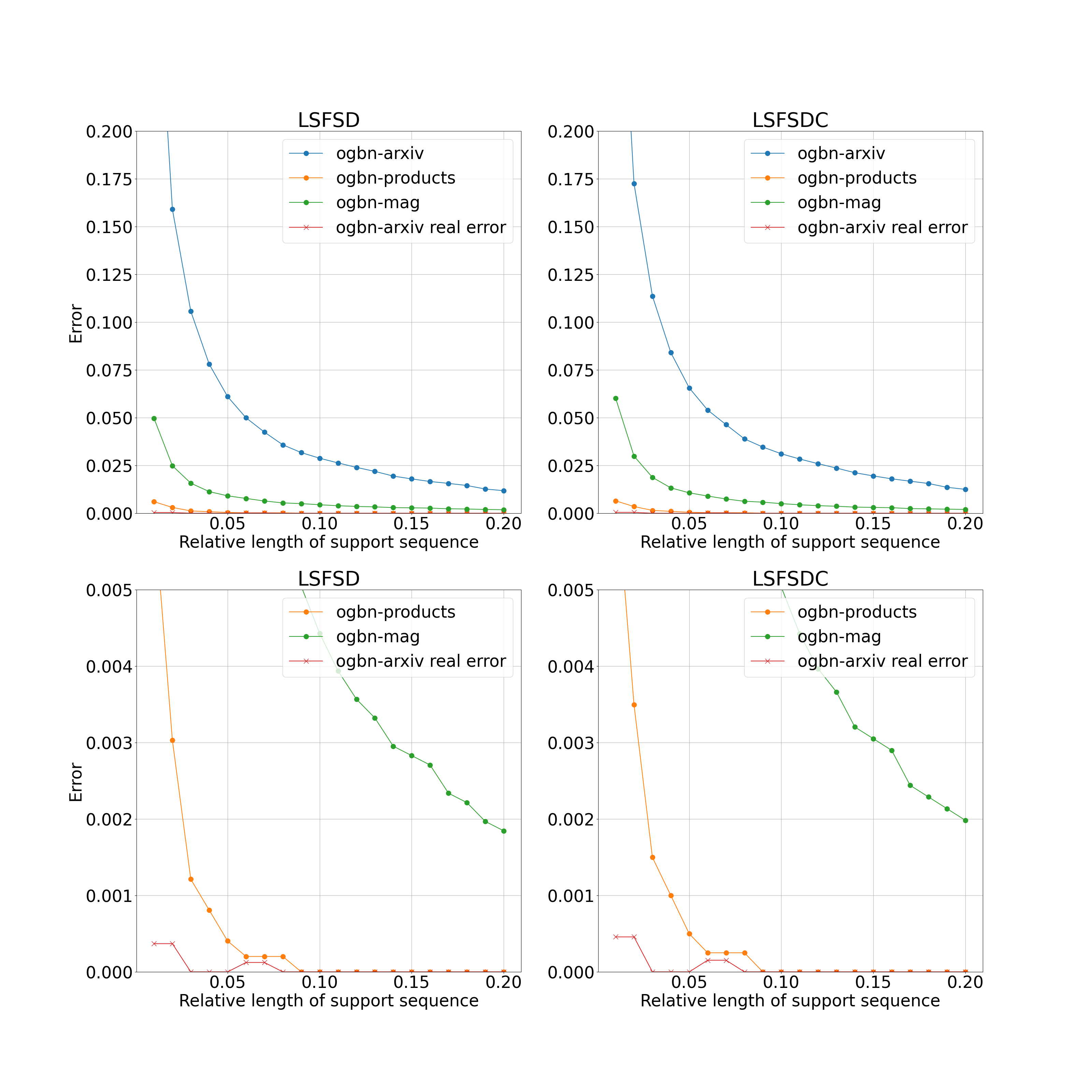}
\caption{The maximal errors $\E(s)$ of all \textbf{ogbn} datasets. On the left, we
display the results without discarding highly correlated features. On the right,
we display the results with. For \textbf{ogbn-arxiv}, we also display the
``real error'' $\E(s)*$. The plots in the second row are zooms into the plots
in the first row.}
  \label{fig:study}
\end{figure}

We now study the influence of the lengths of support sequences on the maximal
error $E(s)$. For this, we generate the support sequence as
in~\cref{sec:exp_gnn} with varying value $l$. To be more detailed, we iterate $l$ through $\{
\lfloor 0.01 * n \rfloor, \lfloor 0.02 *n \rfloor, \dots, \lfloor 0.2 *n
\rfloor\}$, where $n$ is the amount of points in the respective datasets. Here,
if $l = \lfloor r*n \rfloor$, we call $r$ the \emph{relative length of $l$}. Note, that
the final support sequence may have a lower amount of elements in $l$ as we
discard doubled elements. For all three \textbf{ogbn} datasets mentioned above,
we display the maximal errors $\E(s)$ with the procedure mentioned
in~\cref{sec:discarding} and without. Since for \textbf{ogbn-arxiv} the exact
computation of $\Delta(\mathcal{D})_{f}$ is possible, we also compute the
``real'' error $\E(s)^{*}$ for this dataset.

\subsection{Results and Discussion}
The results are depicted in~\cref{fig:study}.
We first note that the behavior of the maximal errors is not fundamentally effected by
the decision for or against discarding strongly correlated features. In both
settings, the curves behave similar. Only the absolute values tend to be lower
when no discarding is done. Hence, the following observation and arguments hold
for both cases.

For the larger datasets, \textbf{ogbn-products} and \textbf{ogbn-mag}, the
maximal error $\E(s)$ is  negligible for comparable small relative support sizes. For
relative support sizes of $0.1$, the maximal mistakes are under $0.01$. For
\textbf{ogbn-arxiv} we have an maximal error of around $0.025$ for this relative
support size and overall higher maximal errors for all relative support lengths.

Comparing $\E(s)^{*}$ and $\E(s)$ for \textbf{ogbn-arxiv} it stands out, that
the maximal error is a strong overestimation of the error $\E(s)^{*}$. As computing
$\E(s)^{*}$ for both other datasets is not feasible, we can not verify whether
this is also the case for them. However, we know because of~\cref{eq:bound}, that
$\E(s)^{*}$ is bound by the maximal error which is already negligible for
\textbf{ogbn-products} and \textbf{ogbn-mag}.

\section{Conclusion and Future Work}
We proposed a novel unsupervised feature selection method that accounts for the
intrinsic dimensionality of datasets. Our approach identifies  those features
that are able to discriminate data points and are thus responsible for taming the
curse of dimensionality. Our experiments provide evidence that intrinsic
dimension-based selection of features is competitive with well established
procedures, occasionally outperforming them. Furthermore, we demonstrated
that sampling techniques can be used to scale our feature
selection method to more than millions of data points. 

We identify as a natural next step for future work scenarios where the
features are not given by coordinate projection, i.e., go beyond columns
in tabular data. The generality of our
modeling of features allows to encode, e.g., edge information for graphs.
Thus, our method can be used for edge-sampling procedures that are important for
training Graph Neural Networks. However, the question on how to encode graph
information via feature functions is open.

While the present work emphasizes on understanding \emph{data} by selecting
features that break the curse of dimensionality,
future work has to tackle as well the problem of understanding the behavior of specific
\emph{model classes}. This helps to further strengthen our understanding on the
interplay between the success of learning and the presence of the curse of dimensionality.

\section*{Acknowledgement}
The authors thank the State of Hesse, Germany for funding this work as part of
the LOEWE Exploration project ``Dimension Curse Detector" under grant
LOEWE/5/A002/519/06.00.003(0007)/E19.

\bibliography{literature}

\end{document}